\documentclass[11pt,letterpaper]{article}
\usepackage[hyperref]{acl2017}
\aclfinalcopy
\usepackage{hyperref}
\usepackage[T1]{fontenc}   
\usepackage[utf8]{inputenc}
\usepackage{color}
\usepackage{graphicx}
\newcommand{\testpercentage}{40.42\%}

\newcommand{\trainpercentage}{41.35\%}

\newcommand\words{W}
\newcommand\cols{\mathcal{C}}
\newcommand\ind{ind}
\newenvironment{myitemize}{
  \begin{itemize}
    \setlength{\itemsep}{0pt}
    \setlength{\parskip}{0pt}
    \setlength{\parsep}{0pt}
  }{\end{itemize}}

\newcommand{\metric}{{\tt metric}}

\newcommand{\dimension}{{\tt dimension}}
\newcommand{\datetime}{{\tt datetime}}
\newcommand{\filter}{{\tt filter}}
\newcommand{\typefont}{\tt}

\newcommand\mlgain{$5.2\%$}

\newcommand{\SORTDIMENSION}{{\typefont SORT\_DIM}}
\newcommand{\SORTMETRIC}{{\typefont SORT\_MET}}
\newcommand{\FIRSTLAST}{{\typefont FIRST\_LAST}}

\newcommand{\LOOKUP}{{\typefont LOOKUP}}
\newcommand{\BEFOREAFTER}{{\typefont BEF\_AFTER}}
\newcommand{\SAMEVALUE}{{\typefont SAME\_VALUE}}
\newcommand{\POSITIONBOTH}{{\typefont POS\_BOTH}}
\newcommand{\AORB}{{\typefont A\_OR\_B}}
\newcommand{\DIFFERENCE}{{\typefont DIFFERENCE}}
\newcommand{\HOWMANY}{{\typefont HOW\_MANY}}
\newcommand{\OTHERTYPE}{{\typefont OTHER\_TYPE}}
\title{Abductive Matching in Question Answering}
\author{Kedar Dhamdhere,
  Kevin S.\ McCurley, 
  Mukund Sundararajan, and 
  Ankur Taly\\
    Google Research\\
    Mountain View, USA}
\begin{document}
\maketitle
\begin{abstract}
  We study question-answering over semi-structured data. We introduce
  a new way to apply the technique of semantic parsing by applying
  machine learning \emph{only} to provide annotations that the system
  infers to be missing; all the other parsing logic is in
  the form of manually authored rules. In effect, the machine learning is used to provide non-syntactic matches, a step that is ill-suited to manual rules.
  The advantage of this approach
  is in its debuggability and in its transparency to the end-user. We
  demonstrate the effectiveness of the approach by achieving
  state-of-the-art performance of \testpercentage\ on a standard
  benchmark dataset over tables from Wikipedia.
\end{abstract}

\section{Introduction}
We investigate the problem of answering questions about
semi-structured data.  More specifically, we consider questions about
tabular data. Individual entries may
represent entities, numeric values or dates, though the list of these
types is not specified a priori.

We illustrate our techniques and measure our performance using the
WikiTableQuestions data set that was first studied by Pasupat and
Liang~\cite{pasupat-liang-2015}. This data set was derived from tables
in Wikipedia articles, and consists of a list of crowdsourced
(question, answer, table) triples.  For instance, one table is about
the movies that the actress Mischa Barton has acted in. The questions
in this data set includes simple factual lookups (``In which
movies was Mischa Barton also a producer?''), or may involve a
composition of several analytic functions (``which was the first year
after 1995 in which Mischa Barton acted in more than 4 movies?'').

Most previous approaches to the problem of question answering have
been based on semantic parsing, machine learning, or mixes of the two.
Rule-based systems and machine learning both have strengths and
weaknesses, and we apply a combination of the two in order to gain
some of the advantages from both.  
\subsection{Our Contributions}\label{sec:contributions}
The approach we take is to apply machine-learning to
providing abductive (speculative) matches when we detect that we are missing an
operand.
As an example, for the question [in what movie was barton
  also the producer?], the terms ``barton'', ``producer'' all have
near exact syntactic matches to various table entities. The term
``movie'' is supposed to match the column called ``title''. This is
the kind of match we use machine learning to discover.

As a consequence, this allows the system to remain {\em transparent}.
By this we mean that the
system is be able to provide information about aspects of the
logic that are speculative, and provide provenance for parts of
the query that generate the answer. This will be
described in more detail in section~\ref{section:transparency}.
We do not use 
machine learning either to score parses as
in~\cite{Liang:2016:LES:2991470.2866568} or~\cite{DBLP:journals/corr/HaugGG17}, or to
solve the problem end-to-end like in~\cite{abadi-neural}.

The advantages of our approach are as follows:
\begin{myitemize}
\item It becomes much easier to debug and iterate on quality
  in the system.
\item It allows the system to be “self-aware” as to when it is
  guessing, and to express doubt in communication back to the user
  (For instance: {\em We think you meant:
    In what [title] was barton also the producer.}) Such human-readable
  justifications were suggested in~\cite{raina2005robust}.
\end{myitemize}

We achieve a \testpercentage\ accuracy on the question answering task
on the WikiTables dataset\cite{pasupat-liang-2015}; this is higher
than the best published result we are aware of; 38.7\%
in~\cite{DBLP:journals/corr/HaugGG17}; see Table~\ref{comparison}.
While our approach achieves
higher overall accuracy, there are some questions that previous
approaches answer correctly but our system does not. We leave a
detailed comparision of the the various methods to future work.
\begin{table}[ht]
  \begin{center}
  \small
  \begin{tabular}{|c|c|}\hline
   {\bf System} & {\bf Test accuracy} \\\hline
    \multicolumn{2}{|c|}{\bf Baselines}\\\hline
    \cite{pasupat-liang-2015} & 37.1\% \\\hline
    \cite{abadi-neural} & 34.2\%  \\\hline
    \cite{abadi-neural} & \\
    Ensemble [15 models] & 37.7\%  \\\hline
    \cite{DBLP:journals/corr/HaugGG17} & 38.7\% \\\hline
\multicolumn{2}{|c|}{\bf Our system}\\\hline
    Without ML-based abduction & 35.22\% \\\hline
    With ML-based abduction & \testpercentage\ \\\hline
  \end{tabular}
  \end{center}
  \caption{Comparison of results}
  \label{comparison}
\end{table}




\section{Prior work}
The problem of factual question answering is by now quite old, but the
formulation of the problem and the approaches depend in part upon the
type of corpus that contains the answers. At one extreme is the case
of a full text corpus in which answers are embedded in 
linguistic prose~\cite{Hirschman:2001:NLQ:973890.973891}. At the other
end of the spectrum is when answers are encoded in fully structured
databases, in which case the problem is cast as a natural language
interface to databases (cf.~\cite{nlp-survey}).

Due to lack of space, we are only able to describe a few of the
previous contributions on question answering.  For
instance, \cite{BCFL13, CY13, BL14} use a semantic parsing approach
for answering questions on an open-end knowledge based like Freebase;
see\cite{Liang:2016:LES:2991470.2866568} for a description of the
approach and a survey of related work.  Yin et al.~\cite{YLLK16}
proposes a neural network that encodes both the query and table using
distributed representations, and passes it through a series of
``executor'' networks to generate the answer. The entire network is
trained using question-answer pairs obtained from a synthetic dataset.
Andreas et al.~\cite{ARDK16} propose a hybrid approach where a neural
network for answering questions is obtained by composing smaller
neural ``modules'' (operators) with the composition layout generated
from a syntactic parse of the question.

The first work on the WikiTableQuestions data set appeared
in~\cite{pasupat-liang-2015}, and used semantic parsing to parse
questions into logical forms, with a machine-learned component to score
the logical forms.  The logical forms are represented in lambda
dependency-based compositional semantics~\cite{LiangDCS13}. The
scoring component is a regression model over features extracted from
the query, table and the logical form. The model is tuned on the
training set.  There is also a semantic function abstraction that
invalidates certain logical forms.  The accuracy that they achieve on
this data set was $37.1\%$.

More recently, the problem has also been tackled
using end to end deep learning~\cite{abadi-neural}. Their approach
derives from the ``Neural Programmer'' work of Neelakantan et al.
~\cite{corr:NeelakantanLS15} wherein the question and the table are
fed as input to a recursive neural network that selects operators
and operands at each step of the recursion. The result of applying the
operator at each step is supplied to the next step.
The best single model achieves an accuracy of
$34.2\%$, while an ensemble of 15 models achieves an accuracy of
$37.7\%$.

Finally, there is the work
of~\cite{DBLP:journals/corr/HaugGG17} that uses a hybrid of the
previous two approaches: It replaces the linear regression model used
to score parses by~\cite{pasupat-liang-2015} with a deep neural
network that is featurized much like the one in~\cite{abadi-neural};
that is, it uses a deep neural network only for scoring, still relying on the
grammar-annotator-semantic function framework to generate candidate
parses. Their best single model achieves an accuracy of $34.8\%$ and
an ensemble of 15 models achieves an accuracy of $38.7\%$.

Finally, we are not the first to have applied abductive reasoning in
NLP applications.  In~\cite{raina2005robust}, the authors applied a
combination of an abductive theorem prover and machine learning
to the task of whether one sentence implies the other.

\section{Our approach}
The overall architecture of the system is
derived from a rule-based sematic parsing system
described in a previous paper~\cite{iui}.
The system described in that paper
is currently used by two Google products, namely
Google Analytics and Google Spreadsheets.

\begin{figure}
  \centering
  \includegraphics[width=0.8\linewidth]{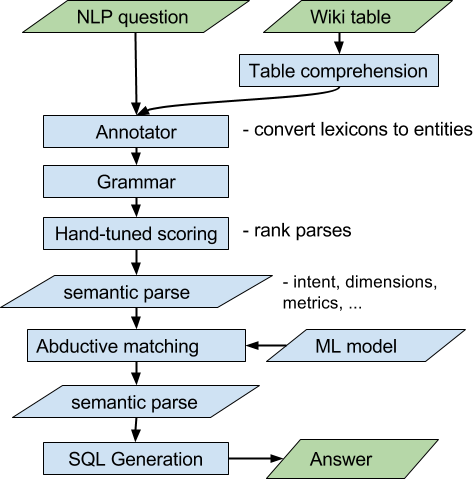}
  \caption{The overall architecture.}
  \label{fig:diagram}
\end{figure}

In this paper we compose the previous system with a machine learning
step to backfill missing pieces that
correspond to unrecognized terms.  The resulting architecture is 
shown in
Figure~\ref{fig:diagram}.
Our main conceptual contribution is
to isolate the use of machine learning to identifying non-obvious term
matches, and we only apply it when abductive reasoning tells us
to. This modularization of the problem raises the question of how to
use the Wikitables training data to learn such matches. The training
data has the form of triples (Question, Table, Answer), so it is not
immediately clear how to generate training data to identify
query-term, table-entity matches. The details of this crucial
traiing process are described in Section~\ref{section:abduction}.

The parsing phase
consists of a standard framework using an annotator and context-free grammar that is designed to
recognize different classes of questions for which there are factual
answers within the tables~(see~\cite{Liang:2016:LES:2991470.2866568}).
The output from the parsing phase is a data structure called
\emph{semantic parse} described in the next section.  Following this
parse phase, we attempt to identify the question type from among a set
of question types~(see Table~\ref{qtype}). Based on this
classification, we enter a phase where we apply abductive
reasoning~(see Section~\ref{section:abduction}) to fill in missing
semantics based on the unmatched terms and question type.

Once the semantic types have been identified, we convert the data structure
into a SQL query on the underlying table. The use of SQL is a convenience
that matches the rectangular structure of the tables, but can easily be
replaced by a query language that is more appropriate to the underlying
data storage.

\subsection{Semantic Types}\label{section:types}
The \textit{semantic parse} contains place holders for the typed
concepts that make up the formal query. The types of these concepts
include {\metric}s (numerical columns), {\dimension}s (string valued
columns), {\filter}s (on dimensions and metrics), and ranges of
{\datetime} values. In addition, the semantic parse contains elements
like sort order, limit, aggregation type, and what type of answer
is expected.
Not all the fields of the semantic parse are filled in for every
query. Further details on the architecture can be found in~\cite{iui}.

\subsection{Table comprehension}
We chose SQL as our formal language for representing logical forms. This
differs somewhat from the approach used
in~\cite{pasupat-liang-2015} where the structured representation was a
more expressive knowledge graph format, with entity normalization
nodes to facilitate the final step of answer
extraction. Rather than using a ``next'' relation,
we simply use a {\tt RowID} column in the table.  The one case in
which these two methods seem to differ is when a cell contains a list
of individual items (e.g., scores of tennis matches). In this case, we
would need to post-process the cells that are needed to extract the
answer, but we found these to be quite rare in the Wikitables data
set.

One of the challenges of unstructured data is that structure is not
explicitly represented in the data, but considerable structure may
still be implicitly represented. As an example, a string like
2005/06/27 will instantly be recognized by a human as a structured
representation for a date without being told it is a date.
There is a limit to how much effort should be put into hand-crafted
parsers for such structures, but we found that the Wikitables data had
several very common features that are explicitly referenced in
questions. 
We
therefore employed several simple parsers to recognize a few
structures, including various date formats, times written as {\tt HH:MM:SS},
numbers formatted in various ways, common units such as km/h, and
scores of sporting events written as {\tt W 21-14}.
In some cases the preprocessing step allows us to split
a column into two or three separate columns.
In case a column contains numeric values, we keep both the original string
values for a column as a \dimension, but may also create 
a separate \metric\ column.
The creation of multiple columns allows
us to easily perform aggregation or differences on numeric values,
but also perform lookups by treating them as string values.
Finally, some tables are already adorned with ``Total'' rows that
contain sums of values above them. In order to allow aggregation over
parts of the table, we separated out these rows when they could be
easily recognized.

\subsection{Annotation and Grammar}\label{sec:annotation}
The output from table comprehension is a \textit{knowledge base} that
maps table entities to (possibly multiple) types. The goal of the
annotator is to use this knowledge base to map phrases in the
user's query with the entities and intent word types. It uses simple
string matching to perform this mapping, augmented by standard
stemming and spell correction. Phrases can have multiple annotations,
and subsequent steps of parsing (such as scoring) perform
disambiguation. The annotator also identifies the
\emph{headword}\footnote{In this paper, a
\emph{headword} is the noun or noun-phrase that succeeds the
question-word, for instance, in our running example, ``movie'' is
headword.} in the
question, and annotates it as a \emph{placeholder} entity.

We use a context-free grammar to parse the annotated query.  The
grammar rules are written in terms of the types. Most of the grammar
rules are ``floating'' in the sense of~\cite{pasupat-liang-2015},
i.e., they ignore the ordering of query terms. We make a few
exceptions, for instance when we parse inequality conditions on
numeric values (``more than 10 wins''). Here we use the sequence of
comparison words, a bound and a metric.

\subsection{Scoring}
We use scoring to produce a
soft ranking among candidate parses. A parse with higher annotation coverage
should be ranked higher. 
The main feature in ranking is the number of annotated question words
for a parse.  Tie-breaking among the candidates with same number of
annotated words is done using features like number of exact matches vs
approximate matches, number of column header matches vs cell
matches. This logic is implemented as a linear model with manually
assigned weights. In the training set, on average our system generated
8.7 candidate parses for each question. By contrast,
in~\cite{pasupat-liang-2015} the number of parses may be exponential in
the number of question terms, but they truncate to~200.

After scoring, we perform the abductive matching step that is our main
contribution. This is described in the next section. After this step,
the semantic parse is turned into an executable SQL query with up to one level
of nesting.  By contrast, \cite{Liang:2016:LES:2991470.2866568} uses a lambda
DCS abstraction, which allows for unlimited composition of operators. Our
semantic parse/SQL abstraction restricts the amount of composition
possible, but we achieve quite good results in spite of this
limitation.
Following execution
of the SQL, there is a normalization step to extract the final answer
based on the question type. We extract a list answer if the \emph{headword} in
the question is plural.

\subsection{The Operand Predictor}\label{section:abduction}

A key conceptual contribution of our work is to separate deductive
reasoning from abductive reasoning. Essentially, we can factor our
system into two parts, namely a rule-based grammar-annotator component
that produces a potentially incomplete parse, followed by a
statistical component called an operand predictor that does its best
to fill in the value of missing operands.

The operand predictor is abductive in the sense of
Mooney~\cite{mooney} that defines abductive reasoning as the
``constructing explanations of observed events''. In other words, the
operand predictor \emph{explains} the incomplete, invalid parse by
adding operands that make the new parse valid, in the sense that it
contains all the required operands.  For our running example, the
additional match of the query term ``movie'' to the column 'Title',
constitutes an explanation that
makes the incomplete parse a valid lookup query.

\paragraph{Identifying missing operands:}
This step is purely deductive: Suppose that the
question is: ``in what movie was barton also the producer?''.
The intent word ``what'' implies that the answer is most likely a
cell of the table, i.e., it is a lookup question.  Second, such a
lookup requires the specification of a row and a column.
Third, the word
``producer'' from the question syntactically matches the phrase ``also
producer'' from a cell of the table. The row of this cell specifies
the row of the answer.  Unfortunately, there isn't a simple syntactic
match between a query term and a column heading, so we are missing
information that identifies the column.

\begin{table*}[!htbp]
\centering  \small
  \begin{tabular}{|l|l|l|l}
Question type & Example & Required Operands\\ \hline
\SORTDIMENSION & which movie has the most budget? & Dimension, Metric\\
\SORTMETRIC & What was the highest attendance & Metric \\
\FIRSTLAST & First movie by Tom Cruise & Dimension \\
\BEFOREAFTER & Actor who won before Tom Cruise & Filter \\
\SAMEVALUE & Which city from same state as Boston & Two Dimensions, Filter \\
\POSITIONBOTH & LA and SF are both in which state? & Dimension, Two Filters \\
\AORB & Who has 4 world cup wins, Germany or Brazil? & Two Filters \\
\DIFFERENCE & What is the difference in height between x and y & Metric, Two Filters \\
\HOWMANY & How many cities with ... & Metric \\
\LOOKUP & Location of Boston Celtics game & Dimension, Filter \\
\OTHERTYPE & (a catch-all for cases we have no semantics for) & at least one column\\
  \end{tabular}
  \caption{Question types used in the statistical component.}\label{qtype}
  \label{table:types}
\end{table*}

Our approach was to identify a list of question
types and required operands with each type.
These are listed in Table~\ref{table:types}.
The detection of the question types is entirely rule-based, using
a combination of intent words and syntactically matched entities. Some question
types are determined purely from the intent words, e.g. ``after'' indicates
\BEFOREAFTER\ and ``how many'' suggests \HOWMANY. On the other hand, to detect
\AORB, we look for the existence of two row filters along with an intent word
like ``or''. Both \SORTDIMENSION\ and \SORTMETRIC\ use words that indicate
ordering (e.g. highest, most). To distinguish between the two, we use presence
of a dimension or metric. The logic for type detection assumes that all row
filters and intent words are identified correctly, but does not require all
columns to have been identified.
Questions that don't fall
under these types (e.g. yes/no questions) also get mapped to \OTHERTYPE.

\paragraph{Predicting operand values:}
In our running example, we have detected a row filter (``producer''),
but we are still missing a dimension operand, and we need to predict its
value. The dimension must be one of the column headings in the table,
which, from left to right are: ``Year'', ``Title'', ``Role'' and ``Notes'', and it
turns out that the correct dimension is ``Title''.

In predicting the correct dimension, the number of column headings is
usually quite small (4-10).
There are
probably a number of ways to implement this prediction. For instance,
by examining the contents of the column, and using the semantic web to
infer that every entry in the column is a movie. Our machine learning approach
is described in the next section.

It is worth noting that this abductive process is closely related to
the fact that real-world questions are often under-specified (see
e.g.~\cite{Small:2004:HTA:1220355.1220544}). For example, the
question ``How much traffic did my website receive?'' is lacking a
time range over which to compute the answer. It make sense to assume
some value for this time range while answering the question and
reflect this assumption back to the user, and we used this approach in
our previous work~\cite{iui}.  This situation does not arise in the
WikiTableQuestions data set, but offers further evidence for why 
abduction is important to the process of question answering.

\subsection{Machine Learning for Abduction}

We notice that there is a frequent co-occurrence of certain query terms
and column headings in the data set. For instance, we notice that the
query term ``movie'' occurs in questions against $43$ tables, and $20$
of these contain the column named ``Title''. This suggests that we can learn
associations ``movie'' $\rightarrow$ ``Title'' and apply these generally to the test set.
See Table~\ref{table:correct} for some examples that we learned.
One could also imagine learning a mapping between a query phrase ``did not swim'' and
the cell entry ``DNS''. Unfortunately they don't co-occur in the training data
frequently enough to learn, so we constrained the learning to only learn the mapping
between query terms and column headings.
By learning such a mapping, we are able to achieve a nearly \mlgain\ gain
in test accuracy.

We train a machine learning model that, given a set of query terms that are
either \emph{unmatched} or assigned to a \emph{placeholder} (see
Section~\ref{sec:annotation}), and a list of columns, assigns probabilities
to each column indicating the likelihood of it being the correct guess.
An \emph{unmatched} term in the question is a term that
cannot be matched to an intent word from the grammar or column heading
or cell value in the table or a placeholder entity.

\subsubsection{Training data generation}
The Wikitables training data set provided only the answer as a label for
a question. For the aforementioned task we seek training data for intermediate
annotation steps. We obtain such data from our parses.

Given parses for the questions that have missing operands,
we construct \emph{counter-factual parses} as follows: We try out all
possible columns for the missing operand in the parse, generating
a new parse for each alternative.
We then generate SQL for each of these parses, and evaluate the SQL over the
table comparing the result to the known correct answer. There are three possibilities.
First, it is possible that  none of the SQL queries lead to the correct answer.
This means that we have not detected the intent correctly, or we do not support the
semantics of the question.
Second, it is possible that more than one of these parses produces a correct answer.
This may happen by
accident. For example, in the question ``how many movies did barton act
in?'', it is possible that the number of distinct values in the
``Title'' and the ``Role'' both result in the right answer.
Third, there is exactly one query leading to the correct answer. Our training data
is constructed from these examples.

For each such example, the training data
consists of a triple $\langle \words, \cols, \ind\rangle$ where
$\words$ is a set of query words, $\cols$ is a list of columns
in the table, and $\ind$ is a one-hot vector (has value
$1$ the correct column). Notice that this process of generating
training data is only tractable because there is a closed, small world
of choices among the columns.

\subsubsection{Training set-up}
We use a very simple neural network for training. We embed each term
into a $50$ dimensional embedding space. We construct query embedding
by adding the embeddings of the query terms together, and a column
embedding by adding the embeddings of the column heading terms
together.\footnote{While the embeddings could be combined using a more complex
  architecture such as an LSTM or a CNN, we prefer our simple averaging method
  for its interpretability.}
We then take the dot product of the two and apply a softmax
that produces a prediction for the correct choice. The loss function
is cross-entropy. We split our training data set into $70/30$
train-test split. When we use the model in serving, it 
generates approximately a \mlgain\ gain in accuracy; see Table~\ref{comparison}.
Note also
that simply guessing the left-most string-valued column would give
a third of this gain.
In our running
example, ``Title'' \emph{is} the left-most string-valued column.

\begin{table}[!htbp]
  \centering \small
  \begin{tabular}{|c|c|r}
Terms  & Column  &  Frequency\\\hline
who  &  name  &  114\\
country  &  nation  &  38\\
who  &  player  &  38\\
player  &  name  &  15\\
film  &  title  &  13\\
who  &  opponent  &  12\\
team  &  opponent  &  11\\
year  &  season  &  11\\
episode  &  title  &  10\\
movie  &  title  &  10\\
movie  &  film  &  8\\
competitor  &  name  &  5\\
  \end{tabular}
  \caption{Examples of \emph{correct mappings} from terms to column names.  }\label{table:correct}
  \label{table:types}
\end{table}

\begin{table}[!htbp]
  \centering \small
  \begin{tabular}{|c|c|c}
Terms & Predicted & Correct\\\hline
tier & division & level\\
size & area (mm2) & diagonal (mm)\\
who & name & president served under\\
who & party & senator\\
game & date & \#\\
win & score & outcome\\
  \end{tabular}
  \caption{Examples of \emph{incorrect} mappings from terms to column names.  }\label{table:wrong}
\end{table}

Tables~\ref{table:correct} and~\ref{table:wrong} show the query term to
column heading matches learned by our model. Table~\ref{table:correct} shows examples
where the model's prediction is correct while Table~\ref{table:wrong} shows examples
where the model's prediction is semantically related but leads to an incorrect
answer.  Notice that the correct
matches predicted by the model are non-syntactic. 
Of the original 14,152 training questions, we derived a training set of
only 1,392 examples. Despite this, certain term/column associations
occur frequently enough to allow for learning.

\subsection{Transparency}\label{section:transparency}

One of our main motivations was to deploy a system that offers a high
level of transparency to both the user and the developer. The user benefits
from seeing how their question was interpreted, and the developer
benefits from being able to debug and
iteratively improve the system.


Most of our system consists of hand-written rules that are easy to debug
in isolation, though the interaction between rules can be quite complex.
Most of the complexity in debugging arises from the annotator and
the abductive matching components.
Errors manifest either as a query-term being unmatched to
 any entity or intent word, or by a query term being matched
 incorrectly. Consequently, our debugging information consists of all
 the entity/intent-word annotations we produced, including the list of
 unmatched query terms. For each of these we include the
 provenance. The types of provenances include \emph{exact syntactic
   match}, \emph{approximate syntactic match}, \emph{machine-learnt
   abductive match}, or \emph{rule-based abductive match}.  The
 developer can use this information in several ways:

\begin{myitemize}
\item To identify unhandled intents by inspecting the list of
  frequently occurring unmatched terms. For instance, we found the terms
  ``next'' and ``previous'' as frequently unmatched in an earlier version. This
  indicated the need to implement position-based selection operator.
\item To debug the approximate matching logic in the annotator and
  the abductively added machine-learning matches.
  \end{myitemize}

Though it does not apply to the WikiTables exercise, we envision
warning the user whenever we use an approximate syntactic match, or
any abductive match: We could preface the response with ``We think the
answer is'', and also identify which query terms if any were used in
the matching. This will help the user decide whether to trust the
response.


\section{Evaluation}\label{sec:eval}
We chose to evaluate our methodology on version 1.0.2 of the
WikiTableQuestions data set~\cite{pasupat-liang-2015}. This dataset
consists of a set of 22,033 (table, question, answer) triples, where
the table is from an HTML page on wikipedia. The data set is
divided into a training set of 14,152 examples and a test set of 4,344
examples, plus an extra set of 3,537 examples that we did not use.
The objective for this data set is to get as many right answers as
possible on the test set, though we are only ever allowed to inspect
the training set.
The questions were sourced through
Mechanical Turk by showing the pages to users and prompting them to
ask a question of a given form. The answers were then collected via
Mechanical Turk by asking other users to answer the questions. In our
evaluation we confined ourselves to using only the CSV form of the
tables, though it is evident from some of the questions in the
training data set that users were shown something more than this. 
See Section 7.2 from~\cite{pasupat-liang-2015} for further details about the dataset.



We have achieved an accuracy of
\trainpercentage\ on the training set, which translates into an accuracy
of \testpercentage\ on the test set (see Table~\ref{comparison}).
We believe that this represents the best
published results on this test set so far.

We expect our system to have different wins and losses from that
of~\cite{pasupat-liang-2015} and ~\cite{abadi-neural}. 
For instance,
~\cite{pasupat-liang-2015} does not handle questions with Yes/No answers.
The end-to-end trained model of \cite{abadi-neural} does not support comparison
operators on derived values and therefore cannot handle questions of type \SAMEVALUE\,
e.g., ``which nation won the same number of bronze medals as peru?''.

Our parse and SQL have bounded ``formula size'' in the terminology
of~\cite{pasupat-liang-2015}. This is possibly
where many of our losses lie. In contrast, their system allows for
arbitrarily long chains of operator composition (``what are the number of movies
that Barton acted in the year after she acted in three movies, two of which were
documentaries?''. On the other hand, our abduction approach (cf. Section~\ref{section:abduction}) lets
us match query terms and column name that aren't close synonyms of each other while
their system is limited in this aspect~\cite[Section 7.5]{pasupat-liang-2015}. Given
the somewhat complementary strengths, it might be interesting to compose our
abduction approach with their system and see what the gain is.

\subsection*{A More Meaningful Evaluation}
While the WikiTableQuestions data set represents a worthwhile evaluation
set to compare the results from different approaches, we believe that it
is not an accurate representation of what a human would expect from a
question-answering system. In particular, there should be a consequence of
giving a wrong answer, but the metrics proposed with the WikiTableQuestions
data set focus only on how many questions are answered correctly. If a human
was asking questions of an analyst or expert, then they would quickly lose
faith in an expert who routinely produced a significant number of answers
that are just wrong. The goal of building reliable question-answering systems
has been previously discussed in~\cite{khani2016unanimity}

As an example, consider the case of yes/no questions. The WikiTableQuestions
training data set has 182 questions of this type, for which 95 (52\%)
have a ``yes'' answer. A system that always answered ``yes'' would therefore
do better than average across all questions, but would clearly result in
an unsatisfying system.
We therefore believe that a metric that penalizes for wrong answers
would better reflect what a real system should deliver.

There are other factors that a real question answering system should address,
such as ambiguity in questions and conversation. In previous work we have
discussed our approach to these issues\cite{iui}.



\section{Summary and Conclusions}
Rule-based systems are transparent, but often not
extensible without significant manual effort.  In contrast, 
machine-learning systems are extensible, but are often not transparent.
We propose an architecture that combines the best of
the two approaches, by using machine-learning only to supply missing
operands. This use of machine-learning is minimal in the sense that
everything that can be easily expressed as rules is expressed as
such. This allows the overall system to be transparent and
debuggable, as discussed in section~\ref{section:transparency}.
While we use machine learning in a limited way, it still has a significant
impact on the accuracy of our system. We expect our architecture with the
use of machine-learnt embeddings within the annotator, combined with a
hand-written grammar, to apply to question-answering on other
corpora.

\bibliographystyle{acl_natbib}
\bibliography{analyza}
\end{document}